# Online and Adaptive Pseudoinverse Solutions for ELM Weights

André van Schaik [a] and Jonathan Tapson [a,*]

[a] *The MARCS Institute, University of Western Sydney, Penrith NSW, Australia 2751*

**Abstract**

The ELM method has become widely used for classification and regressions problems as a result of its accuracy, simplicity and ease of use. The solution of the hidden layer weights by means of a matrix pseudoinverse operation is a significant contributor to the utility of the method; however, the conventional calculation of the pseudoinverse by means of a singular value decomposition (SVD) is not always practical for large data sets or for online updates to the solution. In this paper we discuss incremental methods for solving the pseudoinverse which are suitable for ELM. We show that careful choice of methods allows us to optimize for accuracy, ease of computation, or adaptability of the solution.

\* Corresponding author
  *Email address:* j.tapson@uws.edu.au (Jonathan Tapson).

# 1   Introduction

Since its invention by Huang and colleagues [1] the Extreme Learning Machine (ELM) has become widely used as a method of regression or classification for large and complex data sets. The feature that perhaps most distinguishes the standard ELM from other methods is the analytical solution of the hidden layer weights by means of a matrix pseudoinverse operation. This method allows the rapid and accurate determination of the weights without an incremental or approximate learning process, and makes a significant contribution to the ease of use of ELM. However, there are two ways in which the standard method creates difficulties for users, which are related to the conventional computation of the matrix pseudoinverse by means of a singular value decomposition (SVD) or QR decomposition. The first problem is that these methods are intrinsically batch methods, in which it is assumed that all the input data and all the output target values are available to the user, so that one single computation can be performed; and thereafter the computed hidden layer weights will be fixed and unchanging. The second problem is that as the data sets get large, these methods for computing the pseudoinverse become computationally intractable, as they require both a very large working memory and very large matrix multiplications.

A third issue, which is not as critical for most users, is that an ELM with weights computed using a batch method is not adaptive. If the underlying data-generating process is stationary in a statistical sense, this is not a problem, but in most real-world problems there will be changes in the generative process as time goes by, and the optimal solution will therefore change. It might be helpful to have a weights update rule that gave greater value to the more recent data available, and gradually reduced the contribution of older data.

In this paper we approach these problems by focusing on incremental solutions to the pseudoinverse of a matrix, extending the work on online sequential ELM (OS-ELM) presented in [2]. The purpose of the paper is threefold:

1. To describe the application of Greville's method in ELM so as to produce a weights matrix that is identical to the weights matrix produced by SVD.
2. To outline simplifications to Greville's method, developed by the authors, which reduce the computation required in the incremental algorithm with negligible loss of accuracy.
3. To introduce an adaptive version of Greville's method which allows for nonstationary data.

# 2   Greville's Method

The first incremental solutions to the pseudoinverse of a matrix started appearing not long after Penrose's rediscovery of the pseudoinverse concept [3], and were probably a response to the paucity of computational power available at that time. In particular, a set of incremental methods developed by Edward Greville [4], [5] and subsequently transformed in a simpler algorithmic description by Teuvo Kohonen [6] are very suitable for ELM.  Greville was concerned with the use of the pseudoinverse to solve linear regression problems, so his incremental methods fit very well into the ELM framework. We give a general outline of Greville's methods below.

We will start by outlining the structure of a standard ELM network, with an emphasis on online use. We can consider the data to be a multivariate time series without loss of generality, so that at each sampling time we get a new data vector $x_t \in \mathbb{R}^{L \times 1}$ where $t$ is a time or series index. An ELM network with $L$ input neurons and $M$ hidden layer neurons can be modeled by:

$$y_{n,t} = \sum_{j=1}^{M} w_{nj}^{(2)} f\left(\sum_{i=1}^{L} w_{ji}^{(1)} x_{i,t}\right). \tag{1}$$

The input $x_t$ is weighted by the randomly determined interconnecting weights $w_{ji}^{(1)}$ between the input and hidden layer and summed over $L$ to generate the input to the hidden layer. In (1), the superscript indicates the weights layer; $i$ is the input vector index, $j$ is the hidden layer index, and $n$ is the output layer index. The inputs are then subject to a nonlinear transformation, typically of a compressive nature, although in principle a wide variety of nonlinearities might be suitable. This nonlinear function is indicated by $f(\ )$. A typical ELM also contains a bias term for the hidden layer neurons in this function. This can be absorbed in our formulation by the standard method of assuming that one of the inputs $x_i$ is equal to one for all $t$, and the weight from that input to each hidden neuron then represent that neuron's bias value. The output layer, of dimension $N$, produces a linear weighted sum of the hidden layer activations to produce an output vector $y_t \in \mathbb{R}^{N \times 1}$ corresponding to input $x_t$. The solution of the weights $w_{nj}^{(2)}$ involves the pseudoinverse of the matrix of all hidden layer activations up to the current time $t$.

As in a standard ELM, the input weights $w_{ji}^{(1)}$ are fixed to random values, usually with a normal distribution in some sensible range; for most ELMs $M$ is significantly larger than $L$ and cross validation is used to determine the value of $M$ (typically $M=10L$ in our case); and the output weights $w_{nj}^{(2)}$ are solved by linear regression by means of the pseudoinverse operation.

The output of the hidden layer at time $t$ is a vector $a$ with elements $j$ given by:

$$a_{j,t} = f\left(\sum_{i=1}^{L} w_{ji}^{(1)} x_{i,t}\right). \tag{2}$$

The time series can then be written as $A = [a_1 \cdots a_k]$ where $A \in \mathbb{R}^{M \times k}$ in which each column contains the output of the hidden layer at one instant in the series, and the last column contains the most recent instant in the series. Corresponding to this matrix of hidden layer activations, a similar matrix $Y \in \mathbb{R}^{N \times k}$ consisting of the output values $Y = [y_1 \cdots y_k]$ is constructed. The regression problem then consists of finding the set of weights $W \in \mathbb{R}^{N \times M}$ that will minimize the error in:

$$WA = Y. \tag{3}$$

This may be solved analytically by taking the Moore-Penrose pseudoinverse $A^+ \in \mathbb{R}^{k \times M}$ of $A$:

$$W = YA^+. \tag{4}$$

In the standard ELM implementation, $A$ and $Y$ are the complete (static) data sets, and a singular value decomposition suffices to determine $A^+$. Greville's method shows how $A^+$ may be learnt incrementally in a real-world or real-time application as new data becomes available. We adapt Greville's method below to directly provide updates for $W$, since for this regression, we do not explicitly need $A^+$, but only $YA^+$.

Given the input and output data streams $a$ and $y$ respectively for some process sampled $k$ times, when the next set of data $a_k, y_k$ becomes available, we form two matrices partitioned as follows:

$$A_k = [A_{k-1}\ a_k], \tag{5}$$
$$Y_k = [Y_{k-1}\ y_k]. \tag{6}$$

We now want to calculate $W_k$ given $W_{k-1}$, where:

$$W_k = Y_k A_k^+, \quad W_{k-1} = Y_{k-1} A_{k-1}^+. \tag{7}$$

According to Greville's theorem, the pseudoinverse $A^+$ is given by:

$$A_k^+ = [A_{k-1}\ a_k]^+ = \begin{bmatrix} A_{k-1}^+(I - a_k b_k) \\ b_k^* \end{bmatrix}, \tag{8}$$

where $b_k \in \mathbb{R}^{1 \times M}$ is given by:

$$b_k = \frac{(A_{k-1}^+)^* A_{k-1}^+ a_k}{1 + a_k^*(A_{k-1}^+)^* A_{k-1}^+ a_k}, \tag{9}$$

if

$$c_k \stackrel{\text{def}}{=} (I - A_{k-1} A_{k-1}^+) a_k = 0, \tag{10}$$

where $A^*$ denotes the conjugate transpose (or adjoint) of $A$. In the case where only real valued data is used, the adjoint can be replaced by the transpose.

The condition of equation (10) implies that $a_k$ is in the column space of $A_{k-1}$, i.e., $a_k$ is a linear combination of the previous hidden layer activation vectors $A_{k-1}$. When the number of columns in $A_{k-1}$ is larger than the dimension of $a_k$, i.e., $k > M$, then the vectors $a_k$ are most likely linearly dependent and the condition of equation (10) will hold. This will typically be the case in the online regression defined in equation (3).

To simplify the expression for $b_k$, we can define a symmetric square matrix $\theta_{k-1} \in \mathbb{R}^{M \times M}$:

$$\theta_{k-1} = (A_{k-1}^+)^* A_{k-1}^+ = (A_{k-1} A_{k-1}^*)^+, \tag{11}$$

i.e., $\theta_k$ is the pseudoinverse of the correlation matrix of the hidden layer activation. Equation (9) can then be rewritten as:

$$b_k = \frac{\theta_{k-1} a_k}{1 + a_k^* \theta_{k-1} a_k}, \quad (12)$$

To update $\theta_k$ when a new pair $a_k, y_k$ is observed, we can write:

$$\theta_k = (A_k A_k^*)^+ = (A_{k-1} A_{k-1}^* + a_k a_k^*)^+. \quad (13)$$

Kovanic [7] shows that when (10) holds the matrix inversion lemma may be applied to the pseudoinverse of (13) to obtain an update rule for $\theta_k$. The matrix inversion lemma states:

$$(D + BCB^*)^{-1} = D^{-1} - \frac{D^{-1} BB^* D^{-1}}{C^{-1} - B^* D^{-1} B}, \quad (14)$$

where $D^{-1}$ and $C^{-1}$ exist and $B$ is a matrix such that $BCB^*$ has the same dimensionality as $D$. Choosing $D = \theta_{k-1}^{-1}, B = a_k,$ and $C = I$, gives:

$$\theta_k = \theta_{k-1} - \frac{\theta_{k-1} a_k a_k^* \theta_{k-1}^*}{1 + a_k^* \theta_{k-1} a_k} = \theta_{k-1} - \theta_{k-1} a_k b_k^*, \quad (15)$$

so $\theta_k$ can be readily computed from $\theta_{k-1}$ and $a_k$.

To update the weights, we use:

$$\begin{aligned}
W_k = Y_k A_k^+ &= [Y_{k-1} \ y_k] \begin{bmatrix} A_{k-1}^+ (I - a_k b_k) \\ b_k \end{bmatrix} \\
&= Y_{k-1} A_{k-1}^+ - Y_{k-1} A_{k-1}^+ a_k b_k + y_k b_k \\
&= Y_{k-1} A_{k-1}^+ + (y_k - Y_{k-1} A_{k-1}^+ a_k) b_k \\
&= W_{k-1} + (y_k - W_{k-1} a_k) b_k,
\end{aligned} \quad (16)$$

which can be readily computed from the new pair $y_k$ and $a_k$, and the past values $W_{k-1}$ and $\theta_{k-1}$. We can thus update $W_{k-1}$ and $\theta_{k-1}$ without ever computing $A_k^+$ explicitly. The above method, the Online PseudoInverse Update Method (OPIUM), produces the exact same weights matrix as SVD, as long as condition (10) is met at each iteration. It is also mathematically equivalent to OS-ELM as presented in [2] when using single input-output vector pairs for each update. It is also possible to not do the update with each new vector pair, but rather collect a set of training vector pairs (i.e. a chunk of data) and do an update for each chunk, as shown in [2].

Initially, however, when $k<M$, equation (10) will not hold and we have according to Greville:

$$b_k^* = c_k^+, \quad (17)$$

if

$$c_k \neq 0, \quad (18)$$

In this case the weight update rule (16) does not change, but the online update for $b_k$ needs to be redefined. Kohonen proposes the introduction of a second auxiliary matrix for this purpose in [6], defined as:

$$\Phi_{k-1} = I - A_{k-1}A_{k-1}^{+}. \tag{19}$$

When $c_k = 0$, $\Phi_k = \Phi_{k-1}$, so no update is made to this matrix. However when $c_k \neq 0$, the updates for the variables become:

$$b_k = \frac{\Phi_{k-1}a_k}{a_k^{*}\Phi_{k-1}^{*}\Phi_{k-1}a_k}, \tag{20}$$

$$\Phi_k = \Phi_{k-1} - \Phi_{k-1}a_k b_k^{*}, \tag{21}$$

$$\theta_k = \theta_{k-1} - \theta_{k-1}a_k b_k^{*} + (1 + a_k^{*}\theta_{k-1}a_k)b_k b_k^{*} - b_k a_k^{*}\theta_{k-1}^{*}, \tag{22}$$

which again can be readily computed from the new $a_k$, and the past values $\Phi_{k-1}$ and $\theta_{k-1}$.

## 3  Simplifications

While Greville's method produces the exact pseudoinverse solution for $A_k$, testing for condition (10) in a real world environment is not straightforward, as noise in the input vectors makes it vanishingly likely to hold. Instead, OPIUM initializes the system of equations such that (10) holds and does not need to be tested. One possibility is to use some portion of the data and the standard batch pseudoinverse to initialize the matrices, as proposed in [2]. An alternative possibility for initialization, useful if not much data is available yet or to avoid an initial batch computation, is to modify the problem slightly and ask for $W_k$ to not only map $a_k$ to $y_k$, but also to map arbitrarily small individual activations of hidden layer neurons ($\epsilon I^{M \times M}$) to the zero matrix $O^{N \times M}$. This gives us:

$$A_0 = \epsilon I^{M \times M}, \tag{23}$$
$$Y_0 = O^{N \times M}, \tag{24}$$

With this initialization, $A_0$ is of rank $M$ and the new vector $a_1$ will thus be in the column space of $A_0$, so that condition (10) is met. From this point onwards $k > M$, and we can use equation (9) for $b_k$ each time.

The derivation of OPIUM in the previous section via the Greville method is how we originally obtained the method. Because Greville's method is not generally well known, we considered it instructive to include it here. It is, however, rather involved, particularly if one is to include the proof of Greville's method, which we have omitted here for brevity. A simpler derivation when we know that condition (10) is met can be obtained, as shown in [2], by noting:

$$W = YA^{+} = YA^{*}(AA^{*})^{-1} = \psi\theta, \tag{25}$$

where $\psi$ is the cross correlation matrix between output vectors and hidden layer activation and $\theta$ is the inverse of the autocorrelation of the hidden layer activation. The pseudoinverse identity $A^+ = A^*(AA^*)^{-1}$ is valid if the rows of $A$ are linearly independent. Each row in $A$ contains the activation of a single hidden layer neuron across all input vectors presented so far. Given the random projections of the input vectors to the hidden neurons, followed by a nonlinear activation function in each neuron, it is reasonably likely that the rows of $A$ are linearly independent. The update rule for $\theta_k$ can again be derived from (13) so that equation (15) can be used. The update rule for $\psi_k$ can be derived straightforwardly from (5) and (6):

$$\psi_k = Y_k A_k^* = [Y_{k-1}\ y_k][A_{k-1}\ a_k]^* = \psi_{k-1} + y_k a_k^* \qquad (26)$$

The weights matrix $W$ can thus be calculated anew at each instant using (25) and update rules (15) and (26). Alternatively, we can insert (15) and (26) into (25), which yields, after multiplying and collecting terms, the same update rule as (16) for $W_k$.

From our experiments, we noticed that the off-diagonal elements of $\theta_k$ were always very small, compared to the elements on the diagonal, and that the diagonal elements all had similar values, within a factor 4 or so. In other words, $\theta_k \approx gI$. This indicates that the activation of a hidden neuron across all input patterns is not (much) correlated with that of the other hidden neurons. This is likely a result of the random projections of the input patterns into the hidden layer, and the nonlinear activation function of the hidden neurons. This leads us to propose a further simplification of OPIUM, in which we simply assume $\theta_k = gI$. In this case, we never need to calculate $\theta_k$ and the algorithm simplifies to:

$$b_k = \frac{a_k}{1/g + a_k^* a_k}, \qquad (27)$$

and

$$W_k = W_{k-1} + (y_k - W_{k-1} a_k) b_k, \qquad (28)$$

This version of the algorithm, OPIUM light, is not as precise, but it is very fast. Note that it is not the same as inserting $\theta_k = gI$ directly into (25), which would lead to $W = g\psi$, and a weight update $W_k = W_{k-1} + g y_k a_k^*$ that does not depend on any error signal.

## 4 Nonstationary data

We have so far assumed that there is a static relation between the input vectors and output vectors. For non-stationary data, condition (10) will not always hold, no matter how carefully we initialize the simulation. We could use Greville's alternative expression for $b_k$, but we again run into the problem that we cannot reliably test (10). In addition, the Greville method gives the first training pair $(y_1, a_1)$ the same weight in determining $W_k$ as the most recent pair $(y_k, a_k)$, which is not appropriate if the mapping is nonstationary. We can include a weighting of the training examples as follows:

$$\psi_k = (2 - \alpha)\psi_{k-1} + \alpha y_k a_k^* \qquad (29)$$

$$\theta_k = ((2 - \alpha)A_{k-1} A_{k-1}^* + \alpha a_k a_k^*)^+. \qquad (30)$$

The new update rules then become:

$$\theta_k = \frac{1}{2-\alpha}(\theta_{k-1} - \theta_{k-1}a_k b_k^*), \tag{31}$$

$$W_k = W_{k-1} + (y_k - W_{k-1}a_k)b_k, \tag{32}$$

with

$$b_k = \frac{\theta_{k-1}a_k}{\frac{2-\alpha}{\alpha} + a_k^* \theta_{k-1}a_k}. \tag{33}$$

If $\alpha = 1$, this version reverts to the stationary version of OPIUM, but for $\alpha > 1$, more weight is given to the more recent training examples. The algorithm effectively applies an IIR filter to both correlation matrices. As a result, it shares with IIR filters the potential for instability in recursion, although we have found this to be very uncommon.

## 5   Results

For the static mapping case, the MNIST database, with 60,000 training patterns and 10,000 test patterns, was classified by an ELM using a 728 neuron input layer, a 7280 neuron hidden layer, and a 10 neuron output layer. A simulation using OPIUM was performed in Python and took approximately 20 hours to process all the training patterns once and test all test patterns, while consuming up to 1.8GB of RAM. The classification error, without any pre-processing, such as de-skewing of the digits, was 2.75% [8]. Using OPIUM light the simulation consumed only 1GB of memory and the same full MNIST database was processed in under 7 minutes on the same computer, resulting in a speed up by a factor 175. The classification error increased to 3.7%, so depending on the application, the increase in speed may well outweigh the reduction in precision. In our experiments, $g$ was arbitrarily set to 1, but simulations with $g$ equal to 100, 4, 0.4, and 0.01 produced very similar results, indicating that the choice of $g$ is not critical. For OPIUM, $\epsilon = 0.001$ was used so that this initialization only had a small effect on the final weights matrix. The code for these simulations is available from the authors' website [9]. To investigate performance and execution time dependence on the number of training samples, we repeated the experiment with only 10,000 training patterns, randomly drawn from the full set and again tested the full 10,000 test patterns. The OPIUM simulation took just under four hours and produced an error rate of 10.3%, while OPIUM light only took just under two minutes for an error rate of 6.7%. Note that the memory consumption for both online algorithms is independent of the number of training patterns.

In order to test the performance for a nonstationary mapping, we simulated OPIUM, OPIUM light, and the dynamic version of OPIUM on a task where the mapping between input and output changes halfway through the task. A network with ten input neurons, one hundred hidden neurons and one output neuron was trained to predict the current sample (at time *t*) of a sine wave with frequency *f* and unit amplitude from ten past input samples of the target time series, taken at *t-10dt, t-20dt, …, t-100dt*, where *dt* is the sample interval. The weights from the input neurons to the hidden neurons were drawn from a uniform distribution between -0.5 and 0.5, and the hidden neuron activation was calculated as the hyperbolic tangent of the sum of its weighted inputs. In our simulation, *dt* = 1ms, and *f* = 10Hz for the first 500ms, and *f* = 20Hz for the second 500ms. For OPIUM light, *g* = 1, and for dynamic OPIUM, $\alpha = 1.003$.

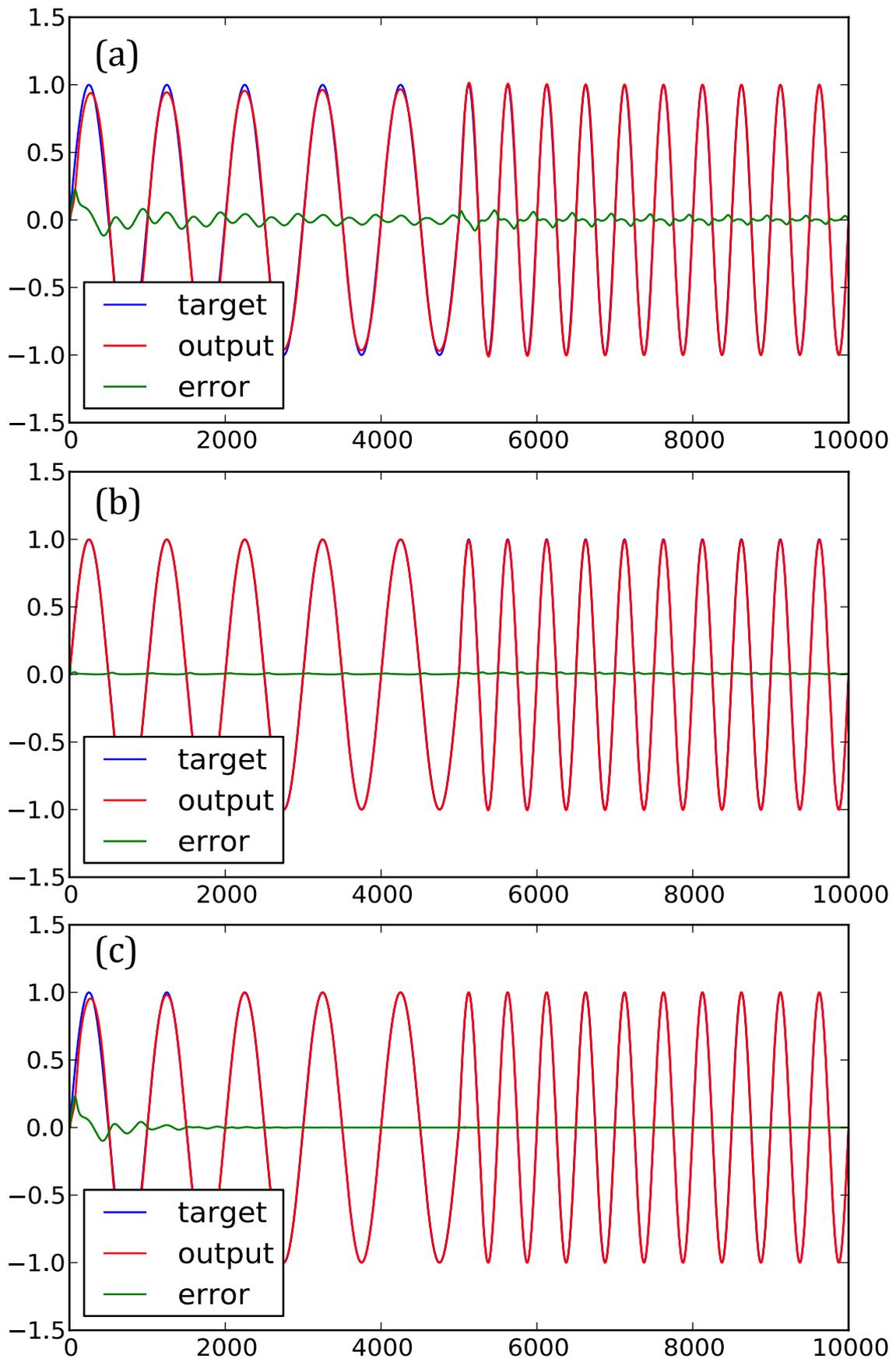

**Figure 1** Example of one simulation run for (a) OPIUM, (b) OPIUM light, and (c) dynamic OPIUM.

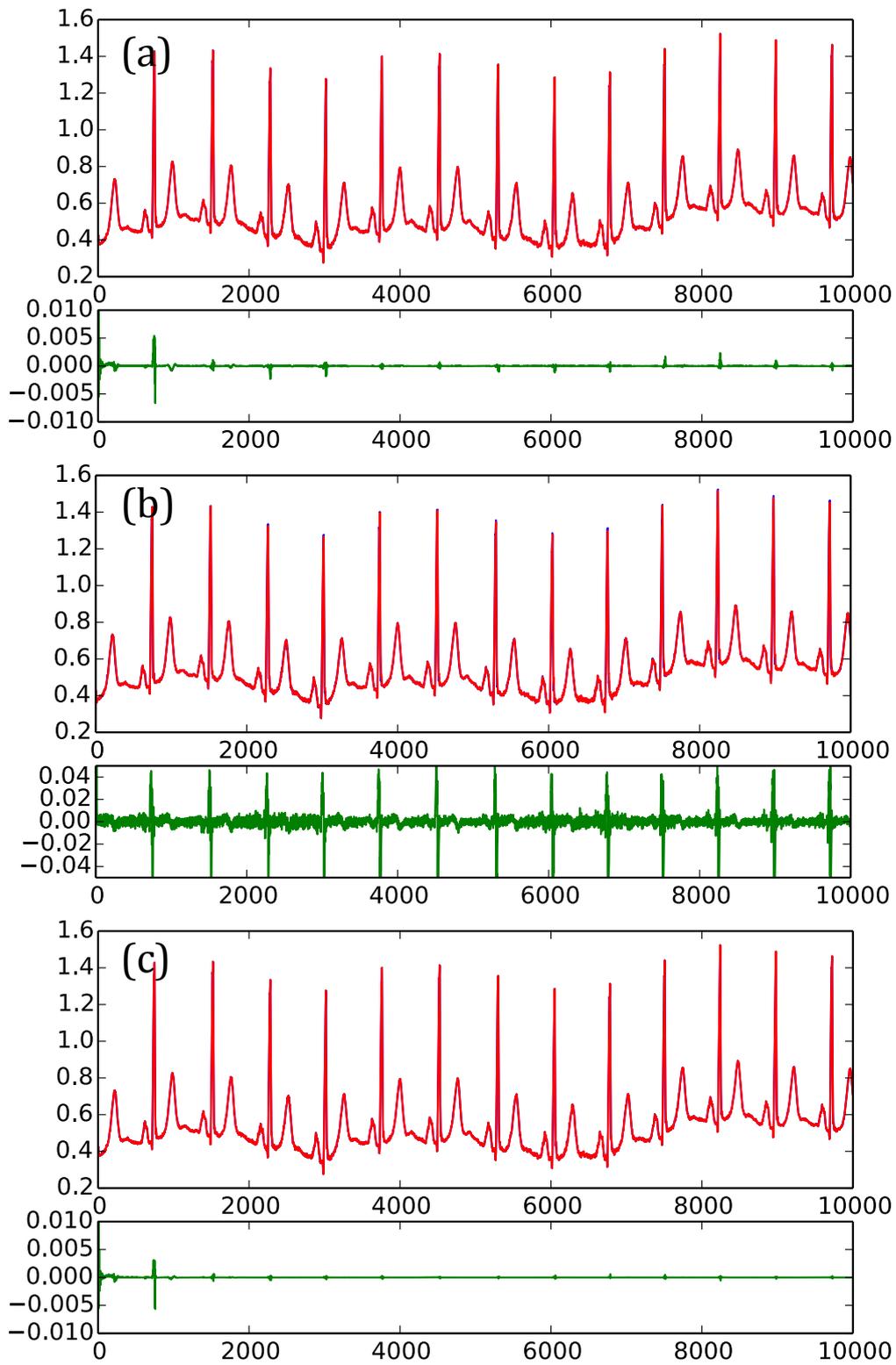

**Figure 2** Example with a recorded ECG signal for (a) OPIUM, (b) OPIUM light, and (c) dynamic OPIUM. Top: target and prediction; bottom: error.

Table 1: RMS error for the final 1000 samples averaged over 100 runs for the simulations of Figure 1

| Algorithm | Average RMS error |
|---|---|
| OPIUM | $1.5 \; 10^{-2}$ |
| OPIUM light | $5.6 \; 10^{-3}$ |
| Dynamic OPIUM | $4.4 \; 10^{-8}$ |

We should point out that in these examples, there are two abrupt transitions of the mapping between inputs and outputs, of which only the second, the change in frequency, is shown. The first change stems from the initialization with (23) & (24), which maps individual activations of hidden layer neurons ($\epsilon I^{M \times M}$) to the zero matrix $0^{N \times M}$. In the above example we have chosen to exaggerate this transition for illustrative purposes by choosing $\epsilon = 10$, whereas normally $\epsilon$ would be chosen to be much smaller than that. From Figure 1 and the results in Table 1 we can see that the standard OPIUM algorithm copes poorly with these changes (Figure 1a), and that the dynamic OPIUM performs very well (Figure 1c). OPIUM light, while not designed for the case where the mapping changes, performs rather well.

We repeated the experiment with an excerpt from a recorded electrocardiogram (ECG) signal to demonstrate the performance with a less predictable signal. In this case, we chose a much smaller $\epsilon = 0.01$, so that the initialization only has a very small effect on the learned weight matrix. Furthermore, we simply used the ten most recent samples of the target time series as the input values for the network to predict the current sample of the target series. All other settings, including the values of the random weights, were the same as in the previous experiment. The results are shown in Figure 2. It can be seen that, with an appropriate initialization, OPIUM performs much better, but that the dynamic version of OPIUM still results in a further improvement.

## 6  Conclusions

In this paper we have described the application of Greville's method to Extreme Learning Machines to obtain an online pseudoinverse update method (OPIUM) that produces a weights matrix identical to that produced by SVD in batch mode. When the condition of equation (10) holds, the derived update method is equivalent to the OS-ELM as derived in [2], when updating the weight matrix every single input-output training pair. Our derivation via Greville's method also provides an update method for the case that equation (10) does not hold. However, in practical implementations, it will be difficult, if not impossible, to test the condition exactly. Therefore, we provide an alternative initialization method after which condition (10) no longer needs to be tested.

We have presented a simplification of OPIUM, OPIUM light, that is very fast, but not as precise. Finally we presented an algorithm designed to cope with a nonstationary mapping between input and output vectors. The inclusion of a free parameter, $\alpha$, allows the user to control the speed of adaptation in this version.